# Soybean pod and seed counting in both outdoor fields and indoor laboratories using unions of deep neural networks

Tianyou Jiang[1], Mingshun Shao[1], Tianyi Zhang[1], Xiaoyu Liu[1], Qun Yu[1,2]

[1] College of Information Science and Engineering, Shandong Agricultural University, Tai'an 271018, China
`s1729041183@gmail.com`

[2] Huanghuaihai Key Laboratory of Smart Agricultural Technology, Ministry of Agriculture and Rural Affairs, Tai'an 271018, China
`yuqun@sdau.edu.cn`

**Abstract.** Automatic counting soybean pods and seeds in outdoor fields allows for rapid yield estimation before harvesting, while indoor laboratory counting offers greater accuracy. Both methods can significantly accelerate the breeding process. However, it remains challenging for accurately counting pods and seeds in outdoor fields, and there are still no accurate enough tools for counting pods and seeds in laboratories. In this study, we developed efficient deep learning models for counting soybean pods and seeds in both outdoor fields and indoor laboratories. For outdoor fields, annotating not only visible seeds but also occluded seeds makes YOLO have the ability to estimate the number of soybean seeds that are occluded. Moreover, we enhanced YOLO architecture by integrating it with HQ-SAM (YOLO-SAM), and domain adaptation techniques (YOLO-DA), to improve model robustness and generalization across soybean images taken in outdoor fields. Testing on soybean images from the outdoor field, we achieved a mean absolute error (MAE) of 6.13 for pod counting and 10.05 for seed counting. For the indoor setting, we utilized Mask-RCNN supplemented with a Swin Transformer module (Mask-RCNN-Swin), models were trained exclusively on synthetic training images generated from a small set of labeled data. This approach resulted in near-perfect accuracy, with an MAE of 1.07 for pod counting and 1.33 for seed counting across actual laboratory images from two distinct studies.

## 1   Introduction

The number of pods per plant and the number of seeds per pod are two important factors for predicting the yield of soybean [1, 2]. Manual counting of pods and seeds is labor-intensive, costly, time-consuming, and prone to errors. Thus, developing effective tools for automatically and accurately counting soybean pods and seeds is significant [3, 4]. Detecting and counting soybean pods and seeds prior to harvest in the field can help rapidly estimate yield. Additionally, during laboratory seed testing, where individual soybean plants are brought into the laboratory and pods are picked, precise counting can be performed. Counting soybean pods and seeds in both outdoor fields and indoor laboratories using computer vision techniques can replace the long-time manual counting and significantly accelerate the soybean breeding process.

In outdoor fields, soybean images are very complex, including complex backgrounds, overly dense plants, and severe soybean pod overlap, making the accurate counting of pods and seeds a formidable challenge. While numerous studies have addressed pod and seed counting, most were conducted under controlled indoor conditions with homogeneous backgrounds and minimal overlap [5]. For instance, Xiang S et al. developed YOLO-Pod to count soybean pods, achieving a Mean Absolute Error (MAE) of 4.184, and Li Y et al. introduced "TCNN-seed", which reported an MAE of 13.21 for seed counting [1, 6]. However, these studies were limited to indoor settings with consistent backgrounds. In contrast, outdoor conditions introduce variability that significantly complicates accurate phenotyping and detection. Addressing these challenges, Zhao J et al. enhanced P2PNet and developed P2PNet-Soy for counting soybean seeds, achieving an MAE of 12.94 [5, 7]. Despite its successes, P2PNet, as a purely point-based framework for joint crowd counting and individual localization, struggles with annotating and predicting soybean seeds that are occluded because it solely annotates and predicts the center points of objects.

In indoor laboratories, soybean pods are picked from the plant during seed testing, which simplifies the counting of pods and seeds when placed directly under a camera. Despite this setup's simplicity, existing methods' accuracy remains insufficient. For example, Li Y et al.'s "TCNN-seed" system only achieved an MAE of 13.21 for seed counting [6]. Furthermore, Yang S et al.'s approach, involving segmentation and seed-per-pod estimation, reached a mean Average Precision (mAP) of only 0.674 at an Intersection over Union (IoU) threshold of 0.5 when tested on real-world images [8]. These results highlight the need for improved accuracy in automatic counting of soybean pods and seeds in the laboratory.

To fill these gaps, we developed distinct strategies for outdoor fields and indoor laboratories. For outdoor fields, we annotated not only visible seeds but also occluded seeds to enable YOLO to estimate the number of soybean seeds that are occluded. Based on YOLO, we propose **a.** YOLO combined with HQ-SAM (YOLO-SAM) to enhance model performance by removing background in the images using detected points. **b.** YOLO combined with domain adaptation (YOLO-DA) to enhance the model's generalization across different environmental conditions and soybean varieties. For indoor laboratories, we utilize Mask-RCNN to count pods, complemented by a Swin Transformer module (Mask-RCNN-Swin) to classify the number of seeds on each segmented pod.

To train and test the efficiency of our methods, comprehensive datasets were built. Models for the outdoor fields were trained with 1010 images (560 from existing studies and 450 captured in this study) and tested on images captured in the outdoor field. We trained and benchmarked standard YOLO, YOLO-SAM, and YOLO-DA, with YOLO-DA performing the best, achieving a mean absolute error (MAE) of 6.13 for pod counting and 10.05 for seed counting. Models for indoor laboratories were trained with 2100 synthetic images and tested with 40 real-world images from two different studies. We use Mask-RCNN to segment pods, complemented by a Swin Transformer module (Mask-RCNN-Swin) to classify the number of seeds on each segmented pod, achieving an MAE of 1.07 for pod counting and 1.33 for seed counting. These results demonstrate the high accuracy and practicality of our proposed methods for counting soybean pods and seeds in both outdoor and indoor environments, which should significantly accelerate yield prediction and soybean breeding processes. The source codes and datasets for soybean pod and seed counting in this study are available at  https://github.com/SkyCol/soybean_phenotyping_platform.

## 2    Related works

### 2.1    2D Object Detection

2D Object detection focuses on predicting category tags and bounding box coordinates for each object within an image. Since the introduction of Regions with CNN (RCNN) in 2014, the field has seen rapid advancements through deep learning technologies [9, 10]. Existing methods are primarily divided into one-stage and two-stage detectors. Among these, Faster-RCNN and YOLO are two of the most notable frameworks in their respective categories [11, 12]. Till now, the Ultralytics team is recognized for their significant contributions to the YOLO series, with YOLOv8 being their latest development. Additionally, Center-Net has gained popularity for its key-point-based approach, which eliminates the need for costly post-processing and Non-Maximum Suppression (NMS), achieving a fully end-to-end detection network [13]. More recently, transformer-based architectures like Detection-Transformer (DETR) also become increasingly prevalent [14, 15].

In a specific subfield of object detection known as crown counting, purely point-based frameworks like P2PNet have been introduced, utilizing point annotations as learning targets and directly outputting points to locate individuals without predicting bounding box coordinates [7]. This approach has been applied effectively in plant phenotyping, for example, Lu H et al. developed Tassel-Net for wheat head detection [16]. Zhao J et al. added k-d tree as a post-process to P2PNet and proposed P2PNet-Soy for soybean seed counting, which showed a significantly improved effect on counting unobstructed soybean seeds in the outdoor field [5]. However, this method is limited to obstructed soybean seed detection, as it relies solely on point annotations.

### 2.2    2D instance Segmentation

While semantic segmentation classifies pixels with semantic labels, instance segmentation further partitions each identifiable object. It extends the scope of semantic segmentation by detecting and delineating each object of interest in the image [17]. Currently, two effective frameworks in this area are Mask-RCNN and YOLOv8-seg (an application of YOLACT) [18, 19]. In recent years, instance segmentation has been widely used in the field of plant phenotyping, helping with higher levels of plant localization and segmentation. For instance, Fan X et al. improved Mask-RCNN to segment and count leaves [20]. Yang S et al. improved Mask-RCNN to segment and count soybean pods and seeds in the laboratory [8]. In this study, we use Mask-RCNN to segment soybean pods in indoor laboratories.

### 2.3   Segment-Anything

The Segment-Anything Model (SAM), introduced in 2023, aims to establish a foundational model for image segmentation capable of zero-shot performance [21]. SAM distinguishes itself by its versatility in segmenting nearly any object through simple prompts such as points, boxes, masks, or text. Moreover, SAM can perform a comprehensive segmentation of all image objects by overlaying a dense layer of points as input prompts. We employed an enhanced version of this model, termed HQ-SAM, to precisely segment individual soybean plants by inputting detected points of soybean pods, thereby improving model performance on predicting soybean images with complex backgrounds [28].

### 2.4   Domain adaptation

Domain adaptation methods are developed to transfer knowledge learned from source domains to target domains which share similar objects yet different data distributions, thus enhance the generalization capability of neural networks across different domains [22, 23, 24]. For object detection, Jaw D W et al. proposed a multidomain object detection framework using feature domain knowledge distillation [25]. By integrating generative adversarial networks, an unsupervised knowledge distillation process and a region-based multiscale discriminator, the proposed framework in their study can extract beneficial features from low-luminance images, outperforming other state-of-the-art approaches in both low- and sufficient-luminance domains. In this study, we use YOLO combined with domain adaptation (YOLO-DA), to help the model better learning on soybean images collected from different studies and enhance the model performance on images captured in outdoor fields.

### 2.5   Soybean pod and seed counting

In the field of plant phenotyping, many studies have extensively utilized object detection and instance segmentation techniques to count soybean pods or seeds. Existing studies predominantly focus on either pods or seeds, conducted in either outdoor fields or indoor laboratory settings.

In outdoor fields, Riera et al. applied a detection-based method to multi-view images for pod counting, though it still suffered from low accuracy [26]. Zhao J et al. improved P2PNet into P2PNet-Soy, which is a point-based localization and counting method using cheap dotted annotations for seed counting [5]. Despite the significant improvements, although it greatly improved the model performance for seed counting, P2PNet-Soy struggled with detecting seeds obstructed by pods or branches due to its characteristic of purely point-based object detection framework. To cope with the complex background of images captured in the field, Mathew J explored the application of a depth camera to remove background using distance information before input-ting the image to YOLO model [27]. This approach, however, depends on additional depth camera sensors. In this study, we aim to overcome these problems through different methods. Annotating both visible seeds and occluded seeds allows YOLO to successfully estimate the number of soybean seeds that are obscured. YOLO-SAM effectively isolates foreground soybean plants using detected points to remove background interferences, eliminating the need for extra hardware such as depth cameras. Meanwhile, YOLO-DA enhances the model's generalization across different soybean varieties and variable environmental conditions.

In indoor laboratories, soybean pod and seed counting are performed using two distinct photographic setups. The first involves photographing the entire soybean plant laid flat on a black cloth background to minimize overlap of its parts. The second method entails photographing only the pods, which are detached from the plant and placed on a similar black backdrop. Due to the clearer image of soybean pods in the second scenario, it is beneficial for accurate counting and we chose this setting in this study. Despite the seemingly straightforward task of automatic counting pods and seeds in laboratories, accuracy remains suboptimal in existing studies under both setups. In the first setting, Xiang S et al.'s YOLO-Pod merely achieved an MAE of 4.184 for pod counting while Li Y et al.'s "TCNN-seed" system merely achieved an MAE of 13.21 for seed counting [1, 6]. In the second setting, Yang S et al. implemented an instance segmentation model trained with synthetic images, which were derived from 28 real-world images. The model in their study achieved a mAP of only 0.674 at an IoU of 0.5 when tested on real-world images [8]. Our study adopts a similar approach to image synthesis but refined the process and improved accuracy in pod and seed counting.

## 3     Materials and Methods

### 3.1    Datasets

Two distinct datasets were built for outdoor and indoor detection of soybean pods and seeds, respectively. In this work, images from different studies were also included and annotated. The statistical information of the datasets is shown in Table 1.

Table 1. Datasets used for training and testing in this study.

| Dataset | Usage | Shooting scene | Source | Amount |
| --- | --- | --- | --- | --- |
| Outdoor | Training | Field | [5] | 150 |
|  | Training | Laboratory | [1] | 300 |
|  | Training | Laboratory | This study | 300 |
|  | Training | Laboratory | This study | 260 |
|  | Evaluation | Field | [5] | 121 |
| Indoor | Training | Synthesis | [8] | 1050 |
|  | Training | Synthesis | This study | 1050 |

| | | | |
|---|---|---|---|
| Evaluation | Synthesis | [8], This study | 700 |
| Evaluation | Laboratory | [8], This study | 40 |

**Dataset for outdoor pod and seed counting.** We collected 271 soybean images captured in the outdoor field from a previous study (see Fig. 1a). To supplement more data for model training. we collected an additional 300 images against a black cloth backdrop in the laboratory from prior work (see Fig. 1b). We also captured 300 more images against a black cloth backdrop (see Fig. 1c), and 275 images with soybeans planted in flowerpots (see Fig. 1d). The soybean images in a, c, and d exhibit significant occlusion, whereas image b has comparatively less occlusion. Subsequently, we annotated these images with detection boxes. During the annotation process, we annotated both obscured and exposed sections of each soybean pod as a complete detection box and assigned a seed count to each box, reflecting the total number of seeds within the pod. This enables our model to estimate both visible and hidden seeds. Finally, images were split into 1010(150+300+300+260) images as the training set and 121 images as the evaluation set, where the evaluation set only includes images taken in the outdoor field. Specifically, during the training process of YOLO-DA, all the 1025 images in the training set were defined as the source domain, and the 150 images from outdoor field were defined as the target domain, to enhance model performance on outdoor-field soybean images.

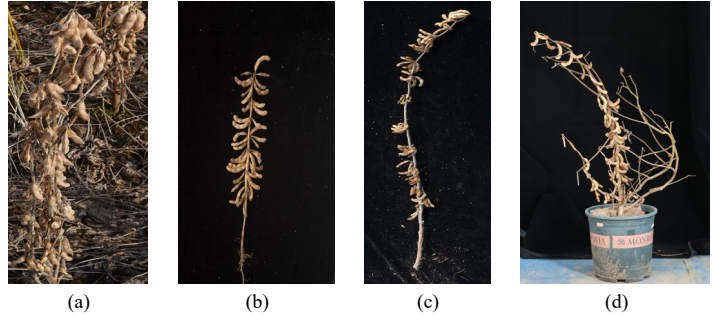

(a)         (b)         (c)         (d)

**Fig. 1.** Samples of soybean images used for outdoor pod and seed counting in this study. (a) image from [1]; (b) images from [5]; (c, d) image captured in this study.

**Dataset for indoor pod and seed counting.** We collected 31 images of soybean pods placed on a black cloth background from a previous study (see Fig. 2a) and added 49 images captured under the same condition in our lab (see Fig. 2b). After labeling for instance segmentation, we selected 40 images (20 from a prior study and 20 newly captured) to generate 2800 synthetic images (see Fig. 2c). This was achieved by randomly selecting, rotating, and placing the masked soybean regions on a pre-prepared black cloth background. The synthetic dataset was split into 2100 training images and 700 evaluation images. Additionally, 40 unaltered images were reserved as a real-world evaluation set to further assess model performance.

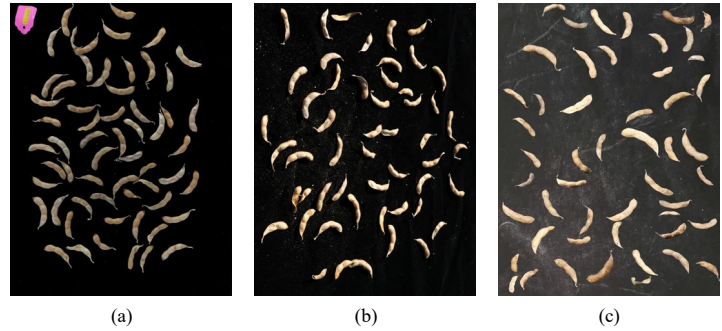

(a)                    (b)                    (c)

**Fig. 2.** Samples of images used for indoor pod and seed counting in this study. (a) real images from [8]; (b) real images captured in this study; (c) synthetic images.

### 3.2  Outdoor model

In outdoor fields, the imaging of soybeans is challenged by severe pod occlusion and overlap, complex backgrounds, and regional varietal differences. While point-based localization and counting models such as P2PNet-Soy deliver reasonable accuracy for visible soybean seeds, they have limitations in detecting soybean seeds that are occluded. To address this, our research adopts the YOLO framework, and the category of the detection boxes were annotated with the overall number of seeds containing both visible and invisible parts of each pod.

Addressing the issue of complex and variable backgrounds, we integrated YOLOv8 with the high-quality segment anything model (HQ-SAM), named as YOLOv8-SAM. This approach isolates soybean plants by utilizing the center points of detected bounding boxes. As illustrated in Fig.3, this method involves three steps: Firstly, YOLO predicts rectangular boxes of soybean pods. Secondly, HQ-SAM extracts soybean plants using the center points of the detected boxes from the first step, achieving isolation by inputting points from the n boxes with the highest confidence. Finally, YOLO detects soybean pods or seeds from the images with black backgrounds and obtain the final counting result.

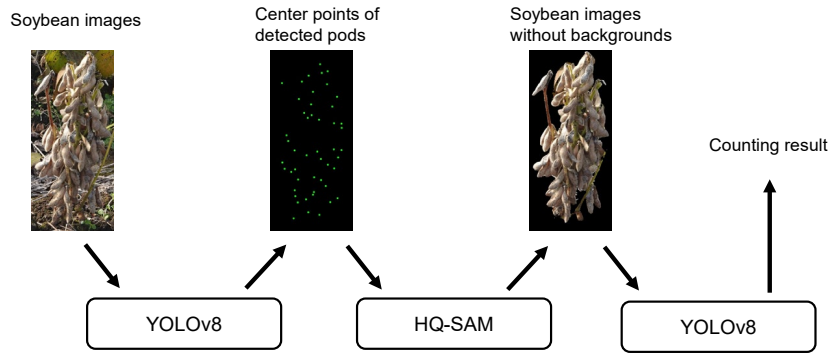

**Fig. 3.** Framework of YOLOv8-SAM in this study.

Furthermore, to cope with diverse soybean varieties and account for varying environmental conditions, we combined YOLOv8 with domain adaptation techniques, named as YOLOv8-DA. This improvement is made through the integration of generative adversarial networks (GANs), and the core idea is from a multi domain object detection framework designed to improve the generalization capability of neural networks across images with varying luminance levels [25]. The framework integrates domain adaptation technology during the training phase, while maintaining the inference process of the original model during the inference phase. In the original paper, domain adaptation was achieved through a multiscale feature fusion discriminator and an unsupervised feature domain knowledge distillation (KD) process. The framework used in this study was simplified and achieved using a GAN with the gradient reversal strategy to replace the KD process. The framework of YOLOv8-DA is illustrated in Fig. 4.

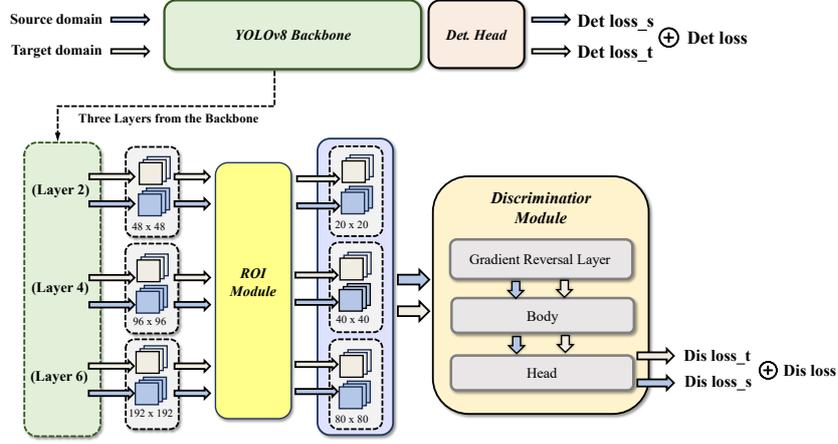

**Fig. 4.** Training process of YOLOv8 (middle) combined with domain adaptation in this study. The loss of source domain and target domain are computed respectively, and the total loss is a sum of the detection loss and discriminator loss.

Given images in the source domain and target domain, the detection loss $L_{Det}$ is computed in the same way as the original YOLO. Feature maps are obtained from three layers in YOLOv8 backbone and are cropped and resized through the ROI module, which allows the discriminator network to concentrate primarily on the object of interest rather than the global image. The Gradient Reversal Layer (GRL) is a crucial component designed to facilitate domain adaptation by reversing the gradients during the backpropagation process [29]. This mechanism enables the feature extractor to learn domain-invariant features effectively. Let $G$ be the feature extractor, $D$ be the domain discriminator, $x_s$ and $x_t$ be the input samples from the source and target domains respectively.

During the forward propagation, the *GRL* directly return x:

$$GRL(x) = x \tag{1}$$

During the back propagation, the *GRL* modifies the gradient as follows:

$$GRL\left(\frac{\partial L}{\partial x}\right) = -\alpha \frac{\partial L}{\partial x} \quad (2)$$

where $\alpha$ is a hyperparameter that controls the strength of the gradient reversal. The discriminator loss $L_{Dis}$ is defined using the hinge loss function:

$$L_{Dis} = E_{x_s \sim S}\left[\max\left(0, 1 - D\left(GRL(G(x_s))\right)\right)\right]$$
$$+ E_{x_t \sim T}\left[\max\left(0, 1 + D\left(GRL(G(x_t))\right)\right)\right] \quad (3)$$

Where $D$ is the discriminator function, $G$ is the feature extractor (layer 2,4,6 and a ROI module here), $x_s$ and $x_t$ are the features from the source domain and the target domain, respectively. $S$ and $T$ represent the source and target domain distributions. This loss encourages the discriminator $D$ to correctly classify the domain of the extracted features while the *GRL* encourages the feature extractor $G$ to confuse the discriminator, leading to domain-invariant features. Then the total loss is:

$$L_{total} = L_{Det} + L_{Dis} \quad (4)$$

In this study, we defined all the 1010 training images containing images captured from both outdoor and indoor as the source domain, and all the 150 outdoor soybean images in the source domain, as the target domain.

### 3.3 Indoor model

When photographing indoor soybean pods against a black cloth background, we use Mask-RCNN to segment pods, complemented by a Swin Transformer module (Mask-RCNN-Swin) to classify the number of seeds on each segmented pod [30]. Given the relatively small and discrete number of seeds per pod, classification effectively addresses the problem with high accuracy, avoiding the uncertainties and potential errors associated with regression method. We conducted benchmark comparisons between Mask-RCNN and YOLOv8-Seg specifically for pod segmentation and opted for Mask-RCNN due to its superior performance over YOLOv8-Seg. Furthermore, we compared the two-step method, Mask-RCNN-Swin, against methods employing either Mask-RCNN or YOLOv8-seg to segment pods with varying seed counts. Our findings indicate that the proposed Mask-RCNN-Swin achieves the highest accuracy, nearing perfection in seed counting.

### 3.4 Training and evaluation settings

In outdoor fields, the box of each pod was labeled with specific categories such as "1spp," "2spp," "3spp," "4spp," where "spp" denotes seeds per pod. Notably when training YOLO-DA, we divided the training set into two domains: all images in the training set as the source domain and the field-captured images within the source domain as the target domain. The standard YOLO model was trained on an NVIDIA RTX A4000 16G GPU, the enhanced YOLO-SAM was trained on an NVIDIA RTX A5000 24G GPU, and the enhanced YOLO-DA was trained on an NVIDIA RTX A100 80G GPU. This differentiation in hardware underscores the diverse resource consumption and computational needs required by each model during their training phases.

In indoor laboratory settings, each soybean pod is meticulously annotated for instance segmentation. For the Mask-RCNN-Swin approach, the initial step involves using Mask-RCNN to segment each pod as a single instance "pod." Subsequently, the number of soybean seeds within each pod is classified, with pods assigned to different categories based on seed count. In contrast, when training the direct methods using Mask-RCNN and YOLOv8-Seg, the instance segmentation annotations are more specific, categorized directly as "1spp," "2spp," "3spp," and "4spp" from the outset. This direct categorization enables these models to segment and simultaneously classify pods based on the number of seeds in a single step. All models in this setting were trained using an NVIDIA RTX A5000 24G GPU.

## 4      Experiments

### 4.1      Outdoor pod and seed counting

The evaluation results of three methods for counting soybean pods and seeds in outdoor fields are shown in Table 2.

**Table 2.** Evaluation results of outdoor pod and seed counting models

| Task | Method | MAE | MAPE |
|---|---|---|---|
| Pod counting | YOLOv8m | 7.03 | 12.75% |
| | YOLOv8m-SAM | 6.99 | 12.61% |
| | YOLOv8m-DA | 6.13 | 9.85% |
| | - | | |
| Seed counting | YOLOv8m | 13.85 | 12.44% |
| | YOLOv8m-SAM | 12.81 | 11.93% |
| | YOLOv8m-DA | 10.05 | 9.42% |

The results showcase high accuracy across all three methods for counting both obstructed and unobstructed soybean pods and seeds. Notably, YOLOv8-DA outperformed the others, achieving an MAE of 6.13 for pod counting and 10.05 for seed counting. In terms of training time and resource requirements, the hierarchy is as follows: YOLOv8m < YOLOv8m-SAM < YOLOv8m-DA. However, when it comes to inference time and resource usage, YOLOv8m-DA is equals to YOLOv8m, and both of them have lower resource consumption compared to YOLO-SAM. A sample of the detection results for occluded pods is illustrated in Fig. 5.

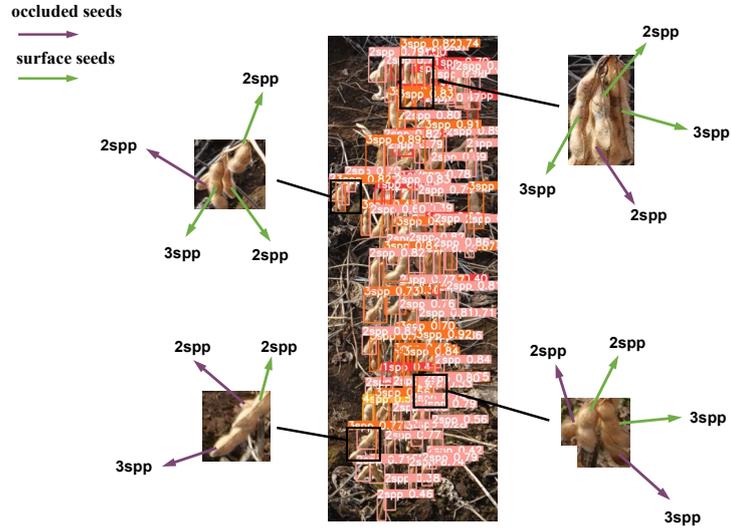

**Fig. 5.** Detection results of occluded seeds.

### 4.2   Indoor pod and seed counting

We use Mask-RCNN to segment each soybean pods, while pod counting was achieved here. After training on the synthetic training set, models were tested on the synthetic evaluation set and real-world evaluation set respectively. Fig. 6 shows the confusion matrix of Mask-RCNN and YOLOv8-seg.

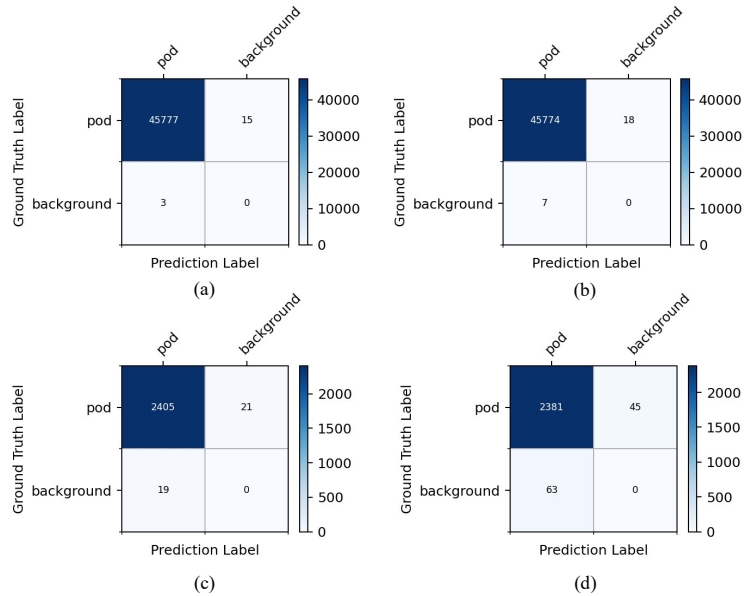

**Fig. 6.** Confusion matrix of soybean pod segmentation. (a) Mask-RCNN on the synthetic evaluation set; (b) YOLOv8-Seg on the synthetic evaluation set; (c) Mask-RCNN on the real evaluation set; (d) YOLOv8-Seg on the real evaluation set.

In the synthetic evaluation set, Mask-RCNN segmented 45777 out of 45792 pods (Fig. 6a), while YOLOv8-Seg segmented 45774 pods (Fig. 6b). In the real-world evaluation set, Mask-RCNN segmented 2405 pods out of 2426 (Fig. 6c), whereas YOLOv8-Seg segmented 2381 pods (Fig. 6d). These results demonstrate that Mask-RCNN exhibits superior performance in pod segmentation compared to YOLOv8-Seg, in both synthetic and real-world evaluation sets.

For the pod counting, we compared the proposed two-step method Mask-RCNN-Swin which uses Mask-RCNN to first segment pods and then calculates the number of seed on each segmented pod, with one-step methods that directly segment pods with different number of seeds. The comparison results on the real-world evaluation set are shown in Table 3.

**Table 3.** Evaluation results of indoor seed counting models

| Method | Counting MAE | Counting MAPE |
| --- | --- | --- |
| Mask-RCNN | 3.57 | 2.32% |
| YOLOv8m-seg | 4.66 | 2.98% |
| Mask-RCNN-Swin | 1.33 | 0.86% |

## 5      Conclusions

In this study, we developed distinct methods for the automatic counting of soybean pods and seeds in both outdoor fields and indoor laboratories. To address the significant occlusion in outdoor soybean images, we labeled the visible and invisible parts of each soybean pod as a whole bounding box and assigned a seed quantity category. Two enhancements to the YOLO framework, namely YOLO-SAM and YOLO-DA, were designed and benchmarked against the original YOLO model in our experiments. Specifically, YOLO-SAM leverages HQ-SAM to remove complex backgrounds in outdoor soybean images by inputting detected points, thereby improving counting accuracy. YOLO-DA employs a domain adaptation training strategy to enhance the model's generalization across different environmental conditions and soybean varieties, demonstrating notable improvements in performance on outdoor soybean images. To further enhance and evaluate the generalization of our models, we plan to capture more outdoor soybean images in the fields.

In practical applications, it is also important to consider the computational consumption. The order of training consumption for the models is YOLO = YOLO-SAM < YOLO-DA, while the inference consumption order is YOLO = YOLO-DA < YOLO-SAM. YOLO-SAM, which uses HQ-SAM for background removal, has a similar training consumption to the original YOLO model due to the additional preprocessing step being relatively lightweight. However, during inference, the background removal process adds extra computational overhead, making YOLO-SAM more resource-intensive compared to YOLO and YOLO-DA. On the other hand, YOLO-DA, which incorporates domain adaptation techniques, requires more computational resources during training to handle the varied environmental conditions and soybean varieties, leading to higher training consumption.

For indoor laboratories, we trained Mask-RCNN using synthetic images to count soybean pods and incorporated a Swin Transformer module to count soybean seeds. By utilizing a small portion of annotated data to generate a large number of synthetic images for training, we achieved high accuracy on real images. This approach allows for rapid model development. The trained model can be applied in laboratory settings to accelerate the processes of seed testing and breeding.

In future work, we aim to develop and integrate more practical applications of soybean phenotyping into our open-source platform, which will further contributing to advancements in this field.

**Disclosure of Interests.** The authors have no competing interests to declare that are relevant to the content of this article.